\documentclass[letterpaper, 10 pt, conference]{ieeeconf}  
\usepackage{floatrow}
\newfloatcommand{capbtabbox}{table}[][\FBwidth]

\usepackage{amssymb}
\usepackage{amsmath}
\usepackage{amsmath,mathtools}
\usepackage{times}
\usepackage{caption}
\usepackage[labelformat=simple]{subcaption}
\usepackage{graphicx}
\usepackage{graphics}
\usepackage{blkarray}
\usepackage{booktabs}
\usepackage{xcolor} 
\usepackage{bbold}
\usepackage{caption}
\usepackage{subcaption}
\usepackage{multirow}
\usepackage{comment}
\usepackage{epstopdf}
\usepackage{tikz}
\usepackage{todonotes}
\graphicspath{{./figures/}}
\usepackage{csquotes}
\usepackage{siunitx}
\usepackage{cite}
\usepackage{epigraph}
\usepackage{empheq}
\usepackage{comment}
\usepackage{cancel}
\usepackage{algorithm} 
\usepackage{algorithmic}  
\usepackage{mathtools}

\usepackage{balance}

\usepackage[linesnumbered,ruled,vlined,algo2e,noend]{algorithm2e}

\SetInd{0.35em}{0.5em}

\usepackage{hyperref}

\SetKwRepeat{Do}{do}{while}

\makeatother

\DeclareCaptionLabelSeparator{periodspace}{.\quad}
\IEEEoverridecommandlockouts                              
\newtheorem{theorem}{Theorem}

\newtheorem{proposition}{Proposition}

\newtheorem{problem}{Problem} 

\newtheorem{definition}{Definition}
\newtheorem{experiment}{Experiment}

\makeatletter
\newcommand*{\rom}[1]{\expandafter\@slowromancap\romannumeral #1@}
\makeatother

\def\eqalignno#1{\let\\=\cr\displ@y \tabskip\@centering
  \halign to\displaywidth{\hfil$\@lign\displaystyle{##}$\tabskip\z@skip
    &$\@lign\displaystyle{{}##}$\hfil\tabskip\@centering
    &\llap{$\@lign##$}\tabskip\z@skip\crcr
    #1\crcr}}
\def\leqalignno#1{\let\\=\cr\displ@y \tabskip\@centering
  \halign to\displaywidth{\hfil$\@lign\displaystyle{##}$\tabskip\z@skip
    &$\@lign\displaystyle{{}##}$\hfil\tabskip\@centering
    &\kern-\displaywidth\rlap{$\@lign##$}\tabskip\displaywidth\crcr
    #1\crcr}}
\begin{document}

\thispagestyle{empty}
\twocolumn
\title{\LARGE \bf
  Act, Perceive, and Plan in Belief Space for Robot Localization}

\author{Michele Colledanchise, Damiano Malafronte, and Lorenzo Natale  
\thanks{The authors are with Humanoid Sensing and Perception group, Istituto Italiano di Tecnologia. Genoa, Italy.
e-mail: {\tt{ michele.colledanchise@iit.it}} \newline This work was carried out in the context of the SCOPE project, which has received funding from the European Union's Horizon 2020 research and innovation programme under grant agreement No 732410, in the form of financial support to third parties of the RobMoSys project.} }

\maketitle
\thispagestyle{empty}
\pagestyle{empty}

\begin{abstract}
In this paper, we outline an interleaved acting and planning technique to rapidly reduce the uncertainty of the estimated robot's pose by perceiving relevant information from the environment, as recognizing an object or asking someone for a direction.

Generally, existing localization approaches rely on low-level geometric features such as points, lines, and planes, while these approaches provide the desired accuracy, they may require time to converge, especially with incorrect initial guesses. In our approach, a task planner computes a sequence of action and perception tasks to actively obtain relevant information from the robot's perception system.


We validate our approach in large state spaces, to show how the approach scales, and in real environments, to show the applicability of our method on real robots. We prove that our approach is sound, probabilistically complete, and tractable in practical cases.

\end{abstract}
\section{Introduction}
\label{sec:introduction}

To navigate, manipulate objects, and perform other tasks a robot requires an accurate estimate of its pose in the environment. Most existing approaches to robot localization, and the related SLAM, rely on low-level geometric features such as points, lines, and planes \cite{thrun2005probabilistic}. While these approaches can provide accurate estimates, they may require long time horizons to converge, especially within environments characterized by ambiguous regions or wherever the localization system receives an incorrect initial guess. 

In this paper, we outline a framework to enhance existing localization techniques by means of a task planning algorithm to actively gather information from the robot's perception system. The proposed framework can estimate the robot's pose within a semantically-annotated map, a map with the expected observations in a given state. We do not aim at surpassing the existing localization and SLAM solutions but rather we aim at improving the existing feature-based localization systems.  We can use the proposed approach to provide an initial guess to the localization system or whenever it has multiple hypotheses valid.

We use semantically-annotated maps since localizing against meaningful landmarks reduces ambiguity \cite{anati2012robot, atanasov2016localization, atanasov2014semantic}, and existing SLAM approaches support the automatic annotation of landmark to speed-up localization (e.g Google Cartographer~\cite{hess2016real}).  Besides, we can exploit the results of the computer vision community, and in particular recent deep architectures for object detection~\cite{redmon2016you,ren2015faster}, to implement perception tasks. Moreover, object detection is robust against changes in viewpoint and noise~\cite{anati2012robot, schonberger2018semantic}, in contrast to local or geometric features, especially given that performance achieved by state-of-the-art architectures for object detection. However, single observations may be ambiguous because identical objects may be found in different locations. For that reason, we propose an \emph{active} approach, in which the robot actively explores the environment to resolve such ambiguity.

Interleaving acting and planning proved to address real world problems such as uncertainty reduction and dynamic environment handling, and provide tractable solutions~\cite{patra2019interleaving,ghallab2014actor,colledanchise2019towards,kaelbling2013integrated, kaelbling2010hierarchical}. Our task planner operates in the \emph{belief space}, a space of probability distribution over possible physical states. Planning in this setting  must account two types of uncertainty: \emph{current state uncertainty} and \emph{next states uncertainty}. To account for the current state uncertainty, in contrast with some approaches that compute the most likely physical state and plan in the physical state, we actively perform actuation and perception tasks to reduce uncertainty  (e.g. move and look for a known landmark). To account for the next states uncertainty, in contrast with some approaches that consider only the most likely outcome of an action, we consider all of them since less likely outcomes can lead to a shorter path in the plan. Whenever the outcome of an action differs from the one considered in the plan, the planner gets re-executed keeping the belief states already searched rather than replanning from scratch. This reduces the replanning times. By planning in belief space, the search algorithm operates in a compact representation of the knowledge about the world, which results in better planning performance. Partially Observable Markov Decision Processes (POMDPs) represent an elegant but impractical formulation of the planning problem in belief space, due to the resulting computational complexity~\cite{russell2016artificial, ingrand2017deliberation, ghallab2016automated}. Moreover, we cannot frame a localization problem with POMDPs as the reward is defined on the physical state or action, hence we have no mean to assign a high reward to low uncertainty.
 
The contribution of this paper comprises:
an interleaved acting and planning framework to rapidly localize the robot in a semantically-annotated map, simulated validation to gather statistically-meaningful data and real robot validation to prove the applicability in the real world. In our example we consider object detection and laser scans as perception capabilities, however our approach can consider any perception capability that gathers information about the current state of the world. In addition, we prove that our approach is sound, probabilistically complete, and scales well in practical cases, comparing it with other solutions.

\section{Related Work}
\label{sec:related}

In this section, we show how researchers exploit object detection to localize robots and the formulation in belief state to find tractable solutions. We also highlight the differences with the proposed approach. We focus on visual-based localization and task planning under uncertainty as no work in the literature addresses our objective of interleaving act an plan with generic information gathering to localize a robot. 
We do not compare our approach with the low-level feature-based localization, as our approach is complementary to those.

Early works on passive visual localization use object detection 
to localize a robot in a prior map annotated with landmarks.
They employ an object detection algorithm to create hypotheses about the robot's pose  within a prior map of landmarks. Then they refine the hypotheses through particle filtering~\cite{anati2012robot}. Recent works include an unified treatment of false positives, false negatives, and data association~\cite{ atanasov2014semantic}. The line of work above still performs object recognition \emph{passively} while the robot moves (i.e. it does not take advantages of the robot capability to actively search for an object). We are interested in planning for actions to actively exploit the robot's perception system to localize itself. Atanasov et al.~\cite{atanasov2016localization} propose  also a so-called \emph{active} object localization, however they assume that the actions sequence is given and is composed only by motion primitives. In our approach we do compute such action sequence and, moreover, we can consider more generic actions than simple motion primitives.

To solve long-horizon task planning in uncertain environments is far beyond the state of the art~\cite{russell2016artificial, kaelbling2013integrated, ghallab2016automated}. Early works rely on the \emph{most-likely-state} approximation where they identify the most likely physical state of the robot, and then act assuming that the robot is on that state. These works are called \emph{determinize-and-plan} approaches. Platt et al.~\cite{Platt-RSS-10} identify a fundamental failing of
these approaches: they never take actions for the explicit purpose of reducing uncertainty, since the planner works on the approximated physical state. However, the work of Platt assumes future observations to be normally distributed about a maximum likelihood and that an observation from a given belief state is always the most probable one. Other works, as~\cite{hadfield2015modular}, extend the classical PDDL planning operators~\cite{ghallab2016automated} with preconditions and effects defined in the belief space to use existing task planners to operate in belief space. They still rely on maximum likelihood observation assumption above.
Kaelbling et al.~\cite{kaelbling2013integrated} proposed an action planning and execution framework to handle uncertainty in robot tasks. They construct task plans in belief space under maximum likelihood observation assumption. Levihn et al.\cite{levihn2013foresight} extended it in terms of smart replanning and reconsideration.

Our approach is different from the ones above in the sense that we define our goal in the belief space directly, we can handle multiple hypothesis, and we do not assume maximum likelihood observation. Planning in the belief space allows efficient overall planning, while removing assuming maximum likelihood observation allows us to take advantages of unlikely effects that lead to short plans.

\section{The Belief Space}
\label{sec:bel}
In this section, we introduce the belief space and we describe how to act, perceive, and plan in this state space.

\begin{figure}[h]
\centering
\begin{subfigure}[t]{0.45\columnwidth}
\centering

\includegraphics[width=0.6\columnwidth]{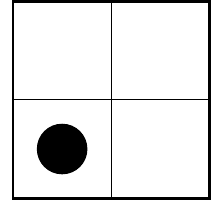}
\caption{Physical state of the grid world with the robot at the bottom left cell.}
\label{INTRO.fig.gridworld-p}
\end{subfigure}
~
\begin{subfigure}[t]{0.45\columnwidth}
\centering

\includegraphics[width=0.6\columnwidth]{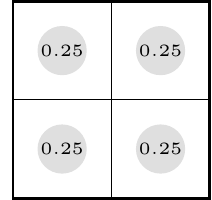}
\caption{Belief state of the grid world with the robot believed to be in any cell with probability 0.25.}
\label{INTRO.fig.gridworld-b}
\end{subfigure}
\caption{Graphical representation of a physical and of a belief state.}
\label{INTRO.fig.gridworld}

\end{figure}

The robot's and world's physical state remain intrinsically non-observable~\cite{thrun2005probabilistic, ghallab2016automated}. The belief space encodes the probability distributions over physical states. For example, consider a grid world with $4$ indexed cells. The physical state $x \in \{0,1,2,3\}$ represents the index of current cell occupied by the robot while the belief state $b \in R_{\geq 0}^{4}$ represent the probability distribution of the robot's position. Figure~\ref{INTRO.fig.gridworld} shows the graphical representation of the physical state $x= 0$ and the belief state $b = \left[0.25, 0.25, 0.25, 0.25\right] $ in a grid world of example.

%


\subsection{Act in Belief Space}
\label{sec:bel:act}

\begin{figure}[h!]
\centering
\begin{subfigure}[b]{0.48\columnwidth}
\centering
\includegraphics[width=\columnwidth]{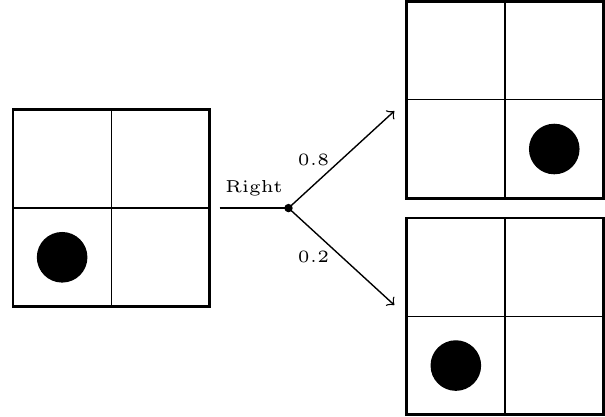}
\caption{In physical space.}
	\end{subfigure}
~
\begin{subfigure}[b]{0.48\columnwidth}
\centering
\includegraphics[width=1\columnwidth]{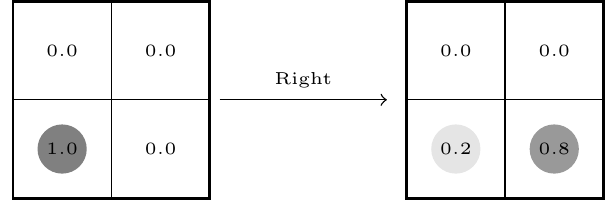}
\vspace{0.25cm}
\caption{In belief space.}
\end{subfigure}

\caption{Graphical representation the non-deterministic effects of the action ``Right" in physical and belief space.}
\label{INTRO.fig.gridworld-b-nd-up}

\end{figure}

By act we refer to those actions that move the physical system from a state to a  \emph{next state}. In general, disturbances make the effect of actions non-deterministic. In the physical space, the non-deterministic effects of an action generate several possible next states. Whereas in belief space, a single next state encodes such non-determinism. Consider the example of the robot in the $2$x$2$ grid world above, where it moves to the desired direction with probability $0.8$ and it does not move with probability $0.2$. Figure~\ref{INTRO.fig.gridworld-b-nd-up} shows the graphical representation of the non-deterministic effects of the action ``Right" in the physical and belief space.

The representation of non-deterministic effects in a single belief state reduces the number of states considered by the search algorithm of the task planner when it generates successors. This results in better performance, as we will describe in Section~\ref{sec:proposed}.
\clearpage
\subsection{Perceive in Belief Space}
\label{sec:bel:sense}

By perceive, we refer to those actions that make a semantic observation on an observable subspace of the world (e.g. detect objects). The effects of those actions depend on the observations from the world, which remain non-deterministic. To perceive reduces uncertainty about the system state and it has no effects on the physical space.
In general, the presence of noise makes the observations imperfect. We can model the noise using the probability of a false-positive $p_{fp}$ and false-negative $p_{fn}$.

\begin{figure}[h]
\centering

\includegraphics[width=0.5\columnwidth]{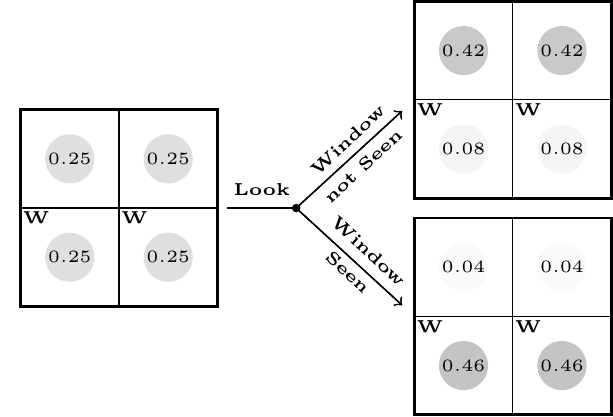}
\vspace*{-1em}

\caption{Belief states reached by the robot after looking for a window, with $p_{fp} = 0.1$ and $p_{fn} = 0.2$. Cells with a window are marked with a ``W".}
\label{INTRO.fig.gridworld-imperfect-sensing-b-path}
\end{figure}

Consider the example of a the robot in the grid world above. To let observation make sense, we add objects to some cells. Cells can have a \emph{Window} and robot can look for it in the current cell. The possible observations for the action \emph{$Look$} are: \emph{$Window Seen$} and \emph{$Window Not Seen$}.  Figure~\ref{INTRO.fig.gridworld-imperfect-sensing-b-path} shows an example of the next belief states with imperfect observations.


%
%
%

\vspace*{-1em}
\subsection{Plan in Belief State}

\begin{figure}[h]
\centering
\includegraphics[width=1.0\columnwidth, , trim=0 0.8cm 0 0, clip]{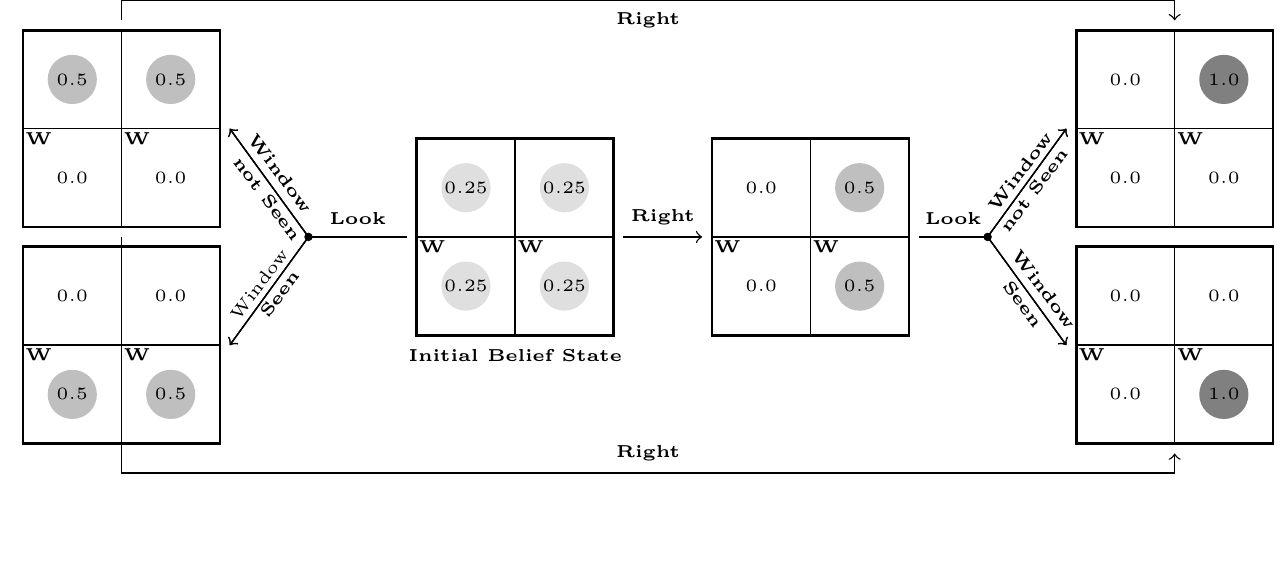}
\caption{Reachable belief states. Self-loop actions are omitted for clarity.}
\label{INTRO.fig.gridworld-sensing-b-graph}
\end{figure}
In real cases, the robot cannot access the physical state but rather only its observable part, from which we can build a belief state.
We can perform forward search from an initial belief state (i.e. the belief state that represents the current probabilistic information of the system's state) to a belief state that represents the desired uncertainty on the physical space.  Consider the grid world of the example above, for simplicity and without loss of generality, we consider perfect observations (i.e. $p_{fn}=p_{fp} = 0$) and only two actions available: the actuation action $Right$ and the sensing action $Look$. Figure~\ref{INTRO.fig.gridworld-sensing-b-graph} shows all the reachable belief states from the initial one. We can see how, starting from the belief state that represents no prior knowledge of the current robot state, any of the plans $Right\to Look$ and $Look\to Right$ moves the belief state to a state that has probability $1.0$ in one physical state and $0.0$ otherwise. Hence, the robot  can execute one of those two plans to localize itself.


\section{Problem Formulation}
\label{sec:problem}

Consider a robot whose transition model, in the general case, is affected by both current and next state uncertainty. The robot has access to a map of the environment containing the observations received for each perception action in each possible robot's physical state in the map. The robot can make semantic measurements, in the general case affected by both false positives and false negatives. Given the above, find a sequence of (actuation and perception) actions that reduces the current state uncertainty of the robot to a given threshold.

\begin{definition}
\label{def:S}
$S$ is a finite set of non-observable physical states. $\Omega$ is a set of observation. 
\end{definition}

An observation is related to a landmark. For example, it can be an object detected (as in \cite{atanasov2016localization, atanasov2014semantic}), the presence of a wall, or the WiFi signal strength (as in \cite{feng2010compressive,sun2014wifi}).

\begin{definition}
$A$ is the set of actions. $A_a \subseteq A$ is the set of actuation actions. $A_p \subseteq A$ is the set of perception actions.
\end{definition}

\begin{definition}
$\gamma:  A \to \mathbb{R}$ is the cost function.
\end{definition}

\begin{definition}
\label{def:transition}
$T_a$ is the $|S|\times |S|$-dimensional one step transition matrix associated to the action $a \in A_a$.
The $i,j-th$ element of such matrix encodes the probability of moving from a physical state $i$ to a state $j$ by performing action $a$ (i.e $P(s_j | s_i,a)$) $\forall i,j \in S, a\in A_a$.
\end{definition}

\begin{definition}
\label{def:observation}

$O_{a}^o$ is the $|S|$-dimensional observation vector associated to the action $a \in A_p$ and observation $o \in \Omega$.
The $i-th$ element of such vector represents the probability of observing $o$ after performing action $a$ in $s_i$ (i.e. $P(o | s,a)$ $\forall o \in \Omega, s_i \in S, a\in A_p$).
\end{definition}


In $T_a$ we can encode actuation errors and disturbance. In $O_a^o$ we can encode false-positives/negatives and the probability to have a landmark in specific position. 

\begin{problem}
\label{problem}
Given the definitions above, let $P(s_i)$ be the probability of being in the physical state $s_i$ and  $\tilde{P} \in [0,1]$ be a desired probability value. Find a  sequence of actions that eventually moves the robot from its initial state $s_i$ to a state such that $P(s_i) \geq \tilde{P}$, if such sequence exists.
\end{problem}

\section{Proposed Solution}
\label{sec:proposed}

In this section, we first provide a set of definitions and then we present the proposed solution.


\begin{definition}
\label{def:belief}
Let $n = |S|$, a belief state is an $n$-dimensional vector $b = \left[\hspace{0.1em} p_0, p_1,\cdots , p_{n-1} \right] $ with $p_i$ the probability of being in state $i \in S$ and
$\sum_{i=0}^{n-1} p_i = 1$.
\end{definition}


%
%


%

\begin{definition}
\label{def:map}

$obs\_map: A_p \to 2^\Omega$ is the \emph{observation map}. It gives the set of observations for a perception action.
\end{definition}

For example, related to the action $Look$, the observation map is a vector that contains, for each observation,  whether an object is detected or not, with possibly a confidence value.

Informally, the proposed approach iteratively computes the belief states and searches for a path to a belief state that satisfies Problem~\ref{problem}.
It does not compute all belief states, as they are infinite in the general case. It starts from a given belief state and computes only the belief states that will be considered by the search algorithm. The search algorithm implements a best-first search algorithm that operates in belief space and has an exit condition based on a property of a belief state. It has as heuristic the entropy of the belief state as we expect the robot to perform more actions to reach the desired uncertainty from a belief state with high entropy.

Formally, Algorithm~\ref{PS.alg.main} describes the proposed approach. 


\vspace{-0.5em}
\subsection{Algorithm Steps in Detail}

\paragraph{Main loop Lines (\ref{alg:main:start}-\ref{alg:main:end})}
The \texttt{Main} function, after initializing some variables, calls the planner from the initial belief (Line~\ref{alg:main:pcall}). In the case of the empty plan, it terminates. Else, the algorithm performs the next action of the plan on the robot and receives the observation  (Line~\ref{alg:main:act}).  
In the case of \emph{None} observation, it performs the next action of the plan. Else, it updates the current belief state applying standard Bayesian inference~\cite{russell2016artificial} (Line~\ref{alg:main:update}). Whenever the updated belief state differs from the expected one, the function calls again the planner  (Line~\ref{alg:main:replan}) giving as input also the graph of nodes already explored (\emph{closed\_list}), else the function performs the next action of the plan. The algorithm terminates if either no solution exists (Line~\ref{alg:main:nosolutions}), the updated belief satisfies the goal condition, or there exists no other action to perform (Line~\ref{alg:main:done}). In the latter case, the goal condition is ensured by the plan (Line~\ref{alg:plan:done}).

\paragraph{Plan (Lines~\ref{alg:plan:start}-\ref{alg:plan:end})}
The \texttt{Plan} function performs best-first search by maintaining a tree of paths originating at the start node and extending those paths one edge at a time until its termination criterion is satisfied (Line~\ref{alg:plan:done}) or there are no other nodes to explore. In detail, a node of the graph  contains the represented belief state ($node.b$) of Definition~\ref{def:belief}, the previous node ($node.from$, if any), the action performed to reach it from the previous node ($node.action$, if any), the cost to reach the node from the initial node ($node.g$), the estimate of the cost required to reach the goal weighted by the probability to reach the node ($node.h$), and the sum of the two ($node.f$). The cost estimate represents a heuristic, and it is defined as the entropy weighted by the probability to reach the node and the maximum belief value (Line~\ref{alg:plan:entropy}).
At each iteration,  the algorithm uses a priority queue ($open\_list$) to determine which of its unexplored paths to extend. It selects the path that minimizes  $node.f\triangleq node.g + node.h$ (Lines~\ref{alg:plan:find:start}-\ref{alg:plan:find:end}). 

The function returns the sequence of actions to solve Problem 1.  The function \texttt{GetPath} (Line~\ref{alg:plan:get}) returns the path which containts actions and the belief states by traversing the graph of nodes backwards from the current node to the initial node.
\paragraph{Next Nodes computations (Lines~\ref{alg:next:start}-\ref{alg:next:end})}
The function \texttt{NextNodes} builds the graph searched in the  function \texttt{Plan}. Given a belief state, it computes all the possible belief states reachable by performing a single action (i.e. the successor nodes of the search algorithm). \begin{algorithm2e}[h]
 \caption{Main Loop}
\label{PS.alg.main}
  \DontPrintSemicolon
  \SetKwFunction{FMain}{Main}
  \SetKwFunction{FPath}{Path}
  \SetKwFunction{FPlan}{Plan}
  \SetKwFunction{FUpdate}{Update}
  \SetKwFunction{FTakeAction}{TakeAction}
  \SetKwFunction{FGetPath}{GetPath}
  \SetKwFunction{FNextNodes}{NextNodes}
  \SetKwProg{Fn}{Function}{:}{}
  \Fn{\FMain{$ start\_belief$}}{ \label{alg:main:start}
 	$closed\_list \gets \emptyset$ ;	$i \gets 0 $; $belief \gets start\_belief$\\
    $path \gets \FPlan(belief, closed\_list)$ \\ \label{alg:main:pcall}
    
    \While{True}{
    \If {$path = NULL$}    
    { \label{alg:main:nosolutions}
		\KwRet NULL
    }

       $obs = \FTakeAction(path[i].action)$\\ \label{alg:main:act}
       
       \If{$path[i].action \in A_p$}
       {
       
         $ belief = \FUpdate( belief, path[i].action, obs)$\\ \label{alg:main:update}
			\If{$max( belief) \geq \bar P $ \textbf{or}  $i=sizeof(path)$}  
			{\label{alg:main:done}
			\KwRet $path$
			}
		\If{$ belief \not = path[i].belief$}
		{\label{alg:main:differ}

		$i \gets 0$      

                   $path \gets \FPlan( belief, closed\_list)$ \\ \label{alg:main:replan}


		}

       }

		$i \gets i +1$ 
		} 
      }\label{alg:main:end}
  
    \SetKwProg{Pn}{Function}{:}{\KwRet}
  \Pn{\FPlan{$belief$, $closed\_list$}}{\label{alg:plan:start}
        $open\_list \gets \emptyset$ \\
        $inital\_node \gets Node(belief)$ \\
        $open\_list.add(b\_initial)$ \\
                    \While{$open\_list \neq \emptyset$}{
                    $candidate\_node \gets open\_list.pop() $\\
                              \For{$open\_node$ in $open\_list$}
                              {\label{alg:plan:find:start}
								\If{$open\_node.f < candidate\_node.f$}
								{
								$candidate\_node \gets open\_node$
								 }  \label{alg:plan:find:end}                            
                               }     
                                $open\_list.remove(front\_node)$\\
                                $closed\_list.add(candidate\_node)$\\ \label{alg:plan:closed}
                                \If{$max(front\_node.b) \geq \bar P$} 
                                {
                                \label{alg:plan:done}
                             \KwRet $\FGetPath(initial\_node, front\_node)$ \label{alg:plan:get}
                                 }
                                
						$ \left[beliefs, acts, probs \right]\gets NextNodes(front\_node.b)$\\

						\For{$i$ $in \left[ 0 \to size\_of(beliefs) \right] $}
						{
						$node \gets Node(beliefs[i])$\\
						$node.act \gets acts[i]$ \\
						$node.from \gets current\_node$ \\

						$node.g \gets  front\_node.g + \gamma(node.act)$\\
						$node.h \gets  \frac{entropy(new\_node.b)}{probs[i]}$\\ \label{alg:plan:entropy}
						$node.f \gets node.g + node.h$\\

						$open\_list.add(node)$\\						
						 }

         }

%
%
\KwRet $NULL$
   }\label{alg:plan:end}

       \Fn{\FNextNodes}{\label{alg:next:start}

        $actions \gets \emptyset$;
        $probabilities \gets \emptyset$;
        $next\_states \gets \emptyset$\\
        \For{$a$ in $A_a$}
        {\label{alg:next:act:start}

        $next\_state \gets T_a \cdot b\_state$\\
        $actions.add(act)$\\
        $probabilities.add(1.0)$\\
        $next\_states.add(norm(next\_state))$\\
         }\label{alg:next:act:end} 
  
       \For{$act$ in $A_p$}       {\label{alg:next:sense:start}
        
            \For{$obs$ in $obs\_map(act)$ }
        		{
       
       $next\_state \gets b\_state \times O(act, obs) $\\

        $actions.add(act)$\\
        $probabilities.add(p(obs | act, b\_state ))$\\
        $next\_states.add(norm(next\_state))$\\

        	}

         }\label{alg:next:sense:end}
        
        \KwRet $\left[next\_states, actions, probs \right]$
   }\label{alg:next:end}
\end{algorithm2e}
 If no such plan exists (i.e. the search explored all the nodes and the termination condition was not satisfied) it returns \emph{NULL}.
 
For each actuation action (Lines~\ref{alg:next:act:start}-\ref{alg:next:act:end}), the function computes the next states. As explained in Section \ref{sec:bel:act}, an actuation action has deterministic effects in belief space. Hence there exists one next belief state for each actuation action. It is computed via the transition matrix $T_a$ of Definition~\ref{def:transition}.

\clearpage

For each perception action  (Lines~\ref{alg:next:sense:start}-\ref{alg:next:sense:end}), the function computes the next states. As explained in Section~\ref{sec:bel:sense}, in general there exists a set of next belief states for each actuation action, computed via the observation map $obs\_map$ of Definition~\ref{def:map} and the observation vector $O$ of Definition~\ref{def:observation}.

The function returns the set of next belief states, the actions and the probability to reach them.

\section{Theoretical Evaluation}
\label{sec:theroetical}
In this section we prove that  Algorithm \ref{PS.alg.main} is sound and probabilistically complete.
%
%

\begin{proposition}
\label{prop:sound}
Algorithm \ref{PS.alg.main} is sound.

\begin{proof}
We need to prove that if a plan exists, the algorithm returns it and if a plan does not exists the algorithm returns \emph{NULL}. The function \texttt{Plan} returns a plan only if it holds the satisfying condition  (Line~\ref{alg:plan:done}). It returns NULL only if all the reachable nodes in the graph are searched and none of them holds the satisfying condition (Line~\ref{alg:plan:end}).
\end{proof}


\end{proposition}

\begin{proposition}
\label{prop:complete}

Algorithm \ref{PS.alg.main} is probabilistically complete.
\begin{proof}
We need to prove that, for each possible initial belief state, the algorithm eventually returns a plan or \emph{NULL}. As mentioned in Section~\ref{sec:proposed}, the belief state represents the probability distribution over physical states and the most probable physical state is given by the maximum entry of the belief state. For each possible belief state the function \texttt{Plan} computes all the reachable belief state using the function \texttt{NextNodes}.
Hence, the graph is built as the algorithm searches on it, and any plan that satisfies Problem $1$ ends in a node of the graph that satisfies $max(node.b) \geq \bar P$, if such node is reached (Line~\ref{alg:plan:done}) the plan returns it. If such plan does not exists, all the possible reachable nodes are eventually searched and, if none satisfies $max(node.b) \geq \bar P$, the plan returns \emph{NULL} (Line~\ref{alg:plan:end}).
\end{proof}


\end{proposition}

\begin{theorem}
Algorithm \ref{PS.alg.main} solves Problem $1$. 

\begin{proof}
By Propositions~\ref{prop:sound} and~\ref{prop:complete}, at each call of the function \texttt{Plan} eventually returns a sequence of actions that is expected to end in a belief state that satisfies the goal condition if any exists. 
Is such plan exists, by Propositions~\ref{prop:sound}, there is a belief state that satisfies the goal condition. The algorithm executes the plan and updates the belief with the function \texttt{Update}. If the updated belief state differs from the expected one in the path, then Algorithm \ref{PS.alg.main} recalls the function \texttt{Plan} from that belief state. Eventually, the Algorithm take actions to explore all possible belief states, hence the one the satisfies the goal condition.
\end{proof}
\end{theorem}

\section{Experiments}
\label{sec:experiments}
We conduct experiments on a simulated environment that allows to collect statistically-significant data and compare our approach with simple solutions. We made the source code available online for reproducibility.\footnote{\url{https://github.com/miccol/ICRA2020Experiments}} We also conducted experiments on a real robot to show the applicability of our approach in the real world. In all experiments, we set the desired probability $\tilde P \triangleq 0.95$. We made available online a video of these experiments and additional ones.\footnote{\url{https://youtu.be/XJnTlrVTZSM}}

\subsection{Simulated Environments}
\label{sec:experiments:sim}
A grid world encodes the simulated environment. Each cell is divided into four triangles to approximate the robot's pose.
Each triangle represents a physical state in $S$ of Definition~\ref{def:S}.

There exists four objects classes. The perception action \emph{Look} computes classes seen by the robot. With this information we automatically construct the observation map of Definition~\ref{def:map} for the sensing action \emph{Look}, that is all the possible permutations of the objects classes seen and not seen. Observations have false-positive and false-negative probabilities set to $0.01$. We automatically compute a semantically-labeled map that contains  the objects seen with the action \emph{Look} from a given state. With this information we automatically construct the observation vector $O$ of Definition~\ref{def:observation}.

There are four actuation actions, two to move backward and forward, and two to rotate $\pi/2$ clockwise and counterclockwise. Each actuation action has a failure probability equal to $0.02$. With this information we automatically construct the transition matrix $T$ of Definition~\ref{def:transition}. We set the cost of actuation actions to $10$ and of perception actions to $1$. 

We ran several experiments with increasing physical state space cardinality on a laptop with and Intel i5-6200U CPU. We ran each experiment for 100 episodes randomizing the number of objects, the position of the objects, and the initial position of the robot.
Tables~\ref{experiments:tab:results1} and~\ref{experiments:tab:results2} collects 
the mean time to plan, the mean number of replans and the mean cost of the plans (whenever one exists) for our approach, compared with two other approaches: one pseudo-random that performs actuation and perception actions in turn and one that uses Algorithm~\ref{PS.alg.main} without heuristic (i.e. we set $h=0$ in Line~\ref{alg:plan:entropy}), which implements a uniform cost search. Compared to the pseudo-random approach, which does not actually plan, our approach computes plans with a cost significantly lower. Compared to the uniform cost search approach, which returns the least-cost plan, our performs better in terms of planning times by one order of magnitude while keeps similar performance in terms of plan cost and number of replans. This results from the heuristic that  makes the function \texttt{Plan} of Algorithm~\ref{PS.alg.main} an informed search algorithm.  
We also see that our approach scales well with the number of states, while the uniform does not handle large state spaces.			Algorithm~\ref{PS.alg.main} keeps track of the belief states explored ($closed\_list$). Table~\ref{experiments:tab:results3} shows how this results in decreasing replanning times.

\vspace{-0.3cm}
\begin{table}[h!]
\begin{tabular}{lcc}
 $|S|$ &  Our  &  Uniform \\
  \cline{2-3} 
\multicolumn{1}{l|}{ $10^2$} & \multicolumn{1}{c|}{$0.0211$s} & \multicolumn{1}{c|}{$0.1424$s}   \\ \cline{2-3} 
\multicolumn{1}{l|}{ $10^3$} & \multicolumn{1}{c|} {$0.7175$s} & \multicolumn{1}{c|}{$13.852$s}  \\ \cline{2-3} 
\multicolumn{1}{l|}{ {$10^4$}} & \multicolumn{1}{c|} {{$6.6937$s}} & \multicolumn{1}{c|}{{---}}  \\ \cline{2-3}
\end{tabular}
\hfill
\begin{tabular}{lcc}
 $|S|$ &  Our  &  Uniform \\
  \cline{2-3} 
\multicolumn{1}{l|}{ $10^2$} & \multicolumn{1}{c|}{$2.58$} & \multicolumn{1}{c|}{$2.02$}   \\ \cline{2-3} 
\multicolumn{1}{l|}{ $10^3$} & \multicolumn{1}{c|} {$4.04$} & \multicolumn{1}{c|}{$3.01$}  \\ \cline{2-3} 
\multicolumn{1}{l|}{ $10^4$} & \multicolumn{1}{c|} {$9.72$} & \multicolumn{1}{c|}{---}  \\ \cline{2-3}
\end{tabular}
  \caption{
  Mean planned times (left)  and number of replans (right). }%
  \label{experiments:tab:results1}
\end{table}
\vspace{-0.7cm}
\begin{table}[h!]
\begin{tabular}{lccc}
 $|S|$ &  Our & Uniform &  Rand \\
  \cline{2-4} 
\multicolumn{1}{l|}{$10^2$} & \multicolumn{1}{c|}{$29.37$} & \multicolumn{1}{c|}{$24.91$} & \multicolumn{1}{c|}{$78.57$}  \\ \cline{2-4} 
\multicolumn{1}{l|}{$10^3$} & \multicolumn{1}{c|} {$ 47.56$} & \multicolumn{1}{c|}{$47.01$} & \multicolumn{1}{c|}{$97.05$} \\ \cline{2-4} 
\multicolumn{1}{l|}{ {$10^4$}} & \multicolumn{1}{c|} {{$61.09$}} & \multicolumn{1}{c|}{{---}} & \multicolumn{1}{c|}{{$300.1$}} \\ \cline{2-4}
\end{tabular}
\hfill
\begin{tabular}{lc}
 & Times  \\ \cline{2-2} 
\multicolumn{1}{l|}{1st} & \multicolumn{1}{c|}{$0.0557$s}  \\ \cline{2-2} 
\multicolumn{1}{l|}{2nd} & \multicolumn{1}{c|}{$0.0161$s}  \\ \cline{2-2} 
\multicolumn{1}{l|}{3rd} & \multicolumn{1}{c|}{$0.0088$s}  \\ \cline{2-2} 
\end{tabular}
  \caption{
  Mean plan cost comparison (left) and first second and third replanning times of our approach of an experiment with $|S|= 10^3$ (right). }%
  \label{experiments:tab:results2}
  \label{experiments:tab:results3}
\end{table}
%

%
%
%

\newpage
\subsection{Real World Experiments}
\label{sec:realex}
We employed an IIT-R1 robot~\cite{parmiggiani2017design} equipped with an 2D camera and a laser scanner. We mapped the environment in a grid with cells divided into triangles. The centers of the cells approximate the robot's positions while the triangles approximate robot's orientations. The robot has to localize itself, without prior information about its pose in the map, in two environments with ambiguous low-level features: a squared kitchen with identical corners and a hallway  with identical walls. There are objects landmarks in known positions.  It is challenging to localize the robot by exploiting only low-level features in these setups due to their symmetries~\cite{Cadena16tro}.  Our approach exploits also object landmarks.

The robot can recognize four object classes: 
\emph{potted plants}, \emph{fire extinguishers}, \emph{chairs}, and \emph{screens}.
The action \emph{Look} gets the object classes seen and not seen from the camera using the YOLO detection architecture~\cite{redmon2016you} trained on samples from the COCO~\cite{lin2014microsoft} and the TUT indoor dataset~\cite{adhikari2018}.

\begin{figure}[t]
\begin{subfigure}{0.175\columnwidth}
\centering
\includegraphics[width=\columnwidth]{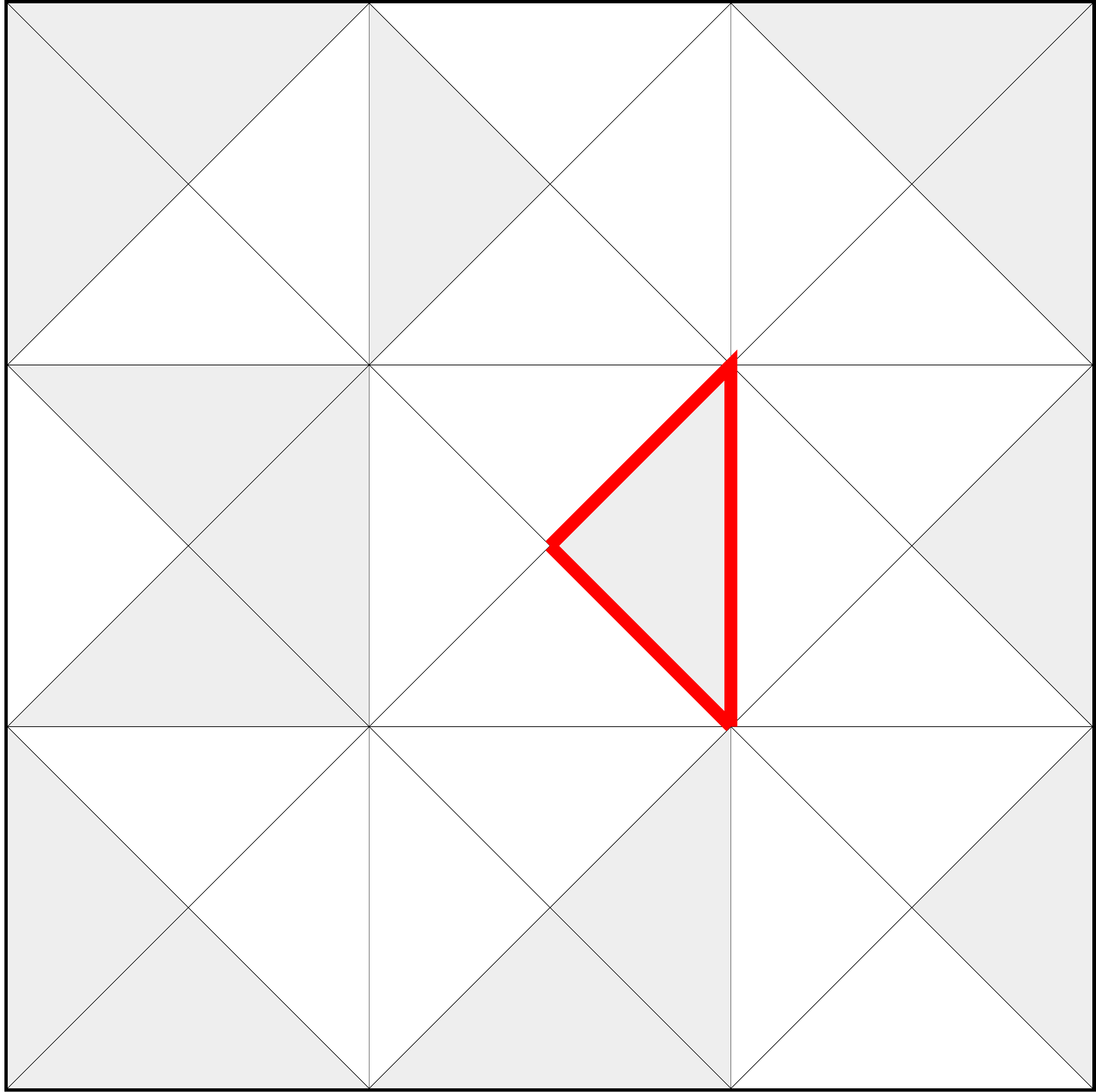}
\caption{Belief. }
\label{experiments:fig:kit:step0:b}
\end{subfigure}
\begin{subfigure}{0.175\columnwidth}
\centering
\includegraphics[width=\columnwidth]{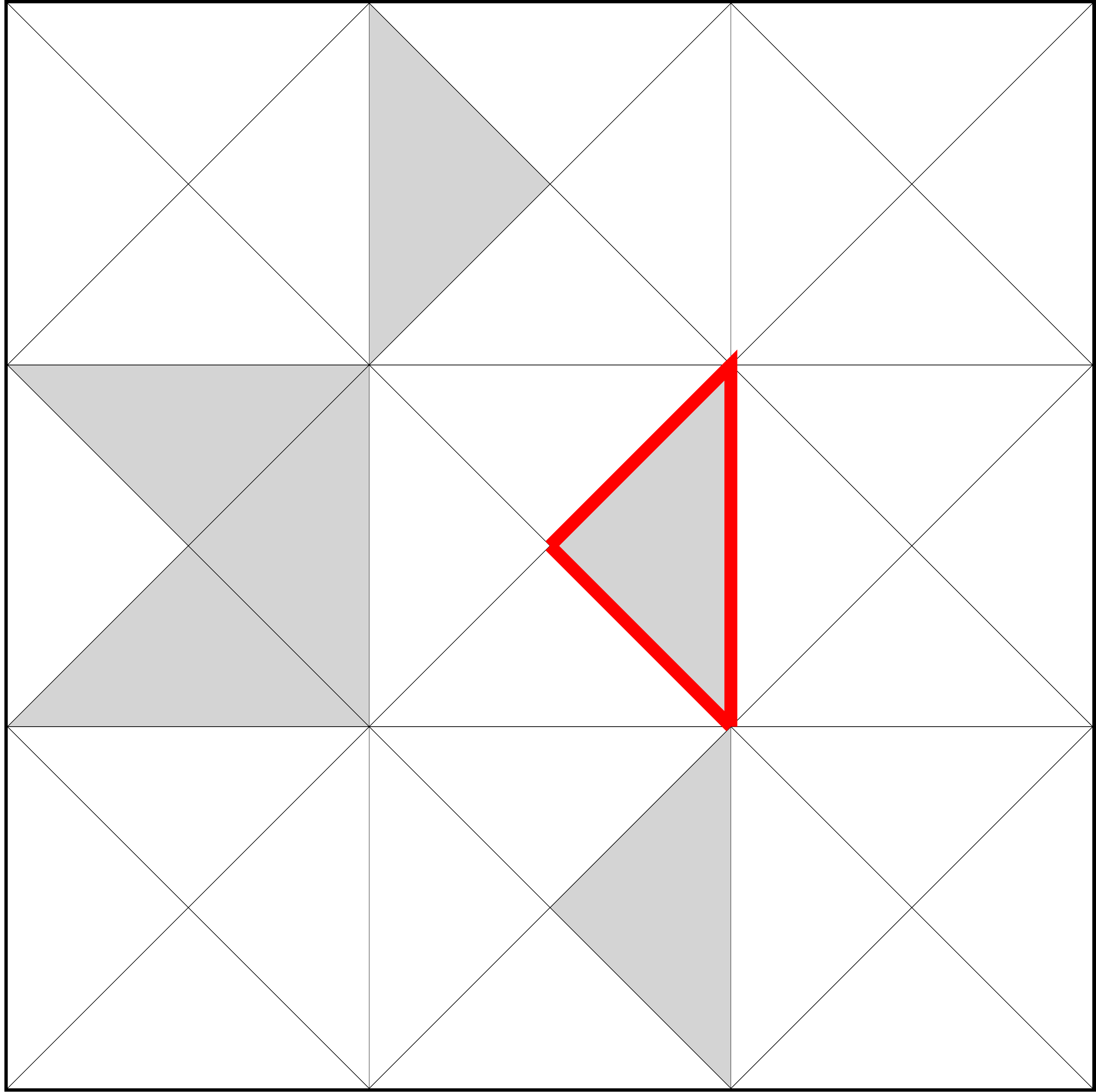}
\caption{Belief. }
\label{experiments:fig:kit:step1:b}
\end{subfigure}
\begin{subfigure}{0.175\columnwidth}
\centering
\includegraphics[width=\columnwidth]{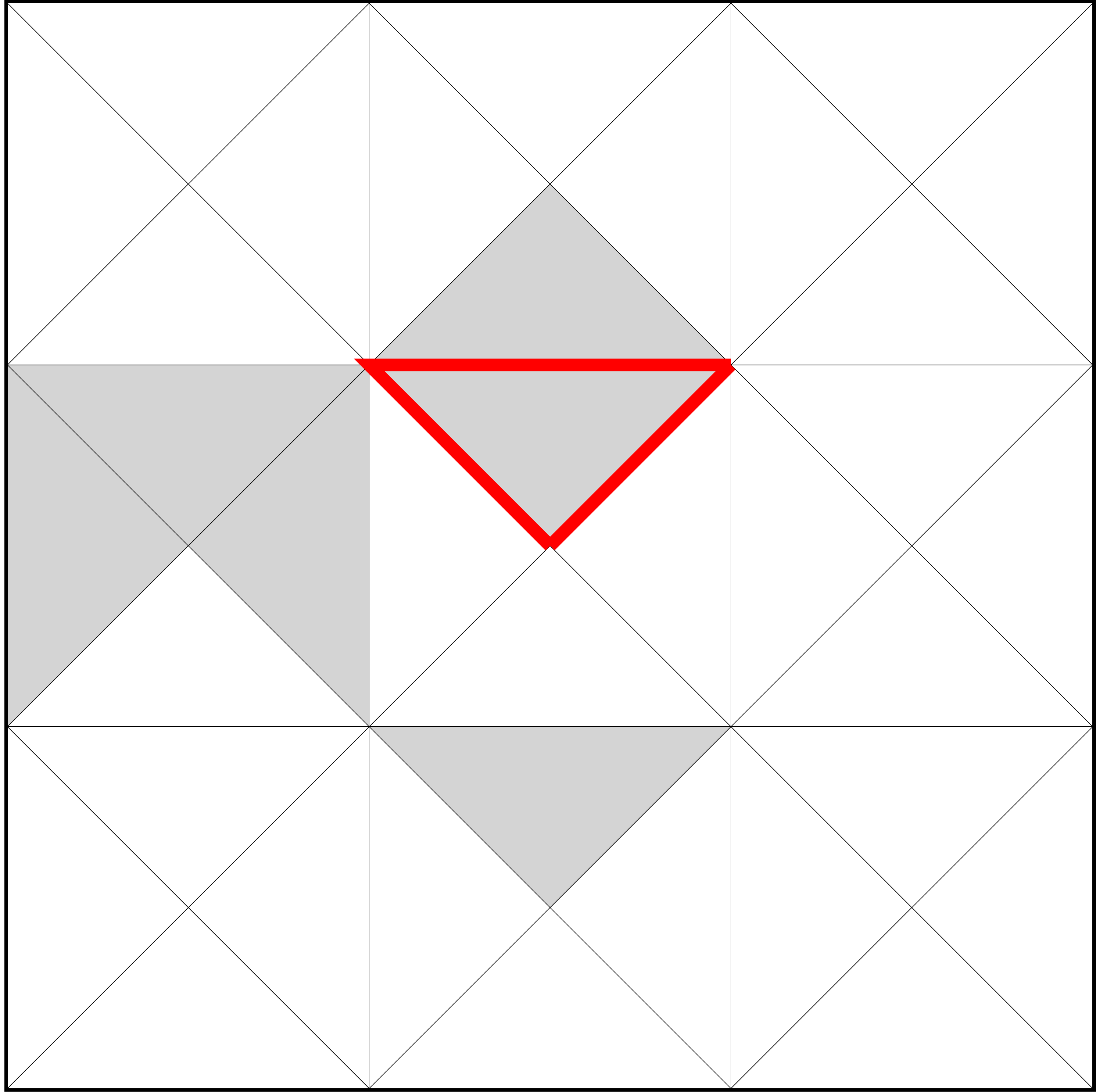}
\caption{Belief. }
\label{experiments:fig:kit:step2:b}
\end{subfigure}
\begin{subfigure}{0.175\columnwidth}
\centering
\includegraphics[width=\columnwidth]{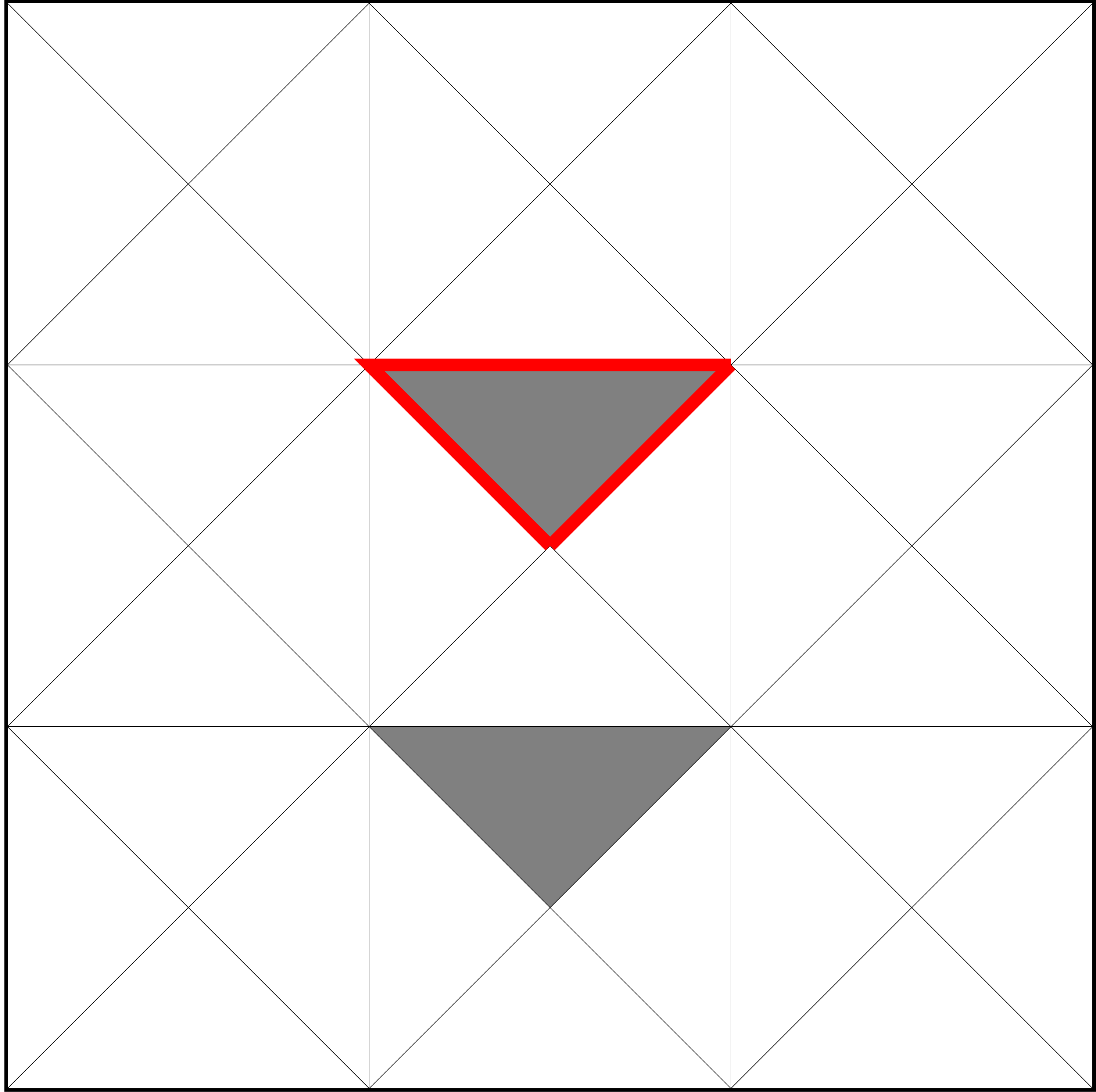}
\caption{Belief. }
\label{experiments:fig:kit:step3:b}
\end{subfigure}
\begin{subfigure}{0.175\columnwidth}
\centering
\includegraphics[width=\columnwidth]{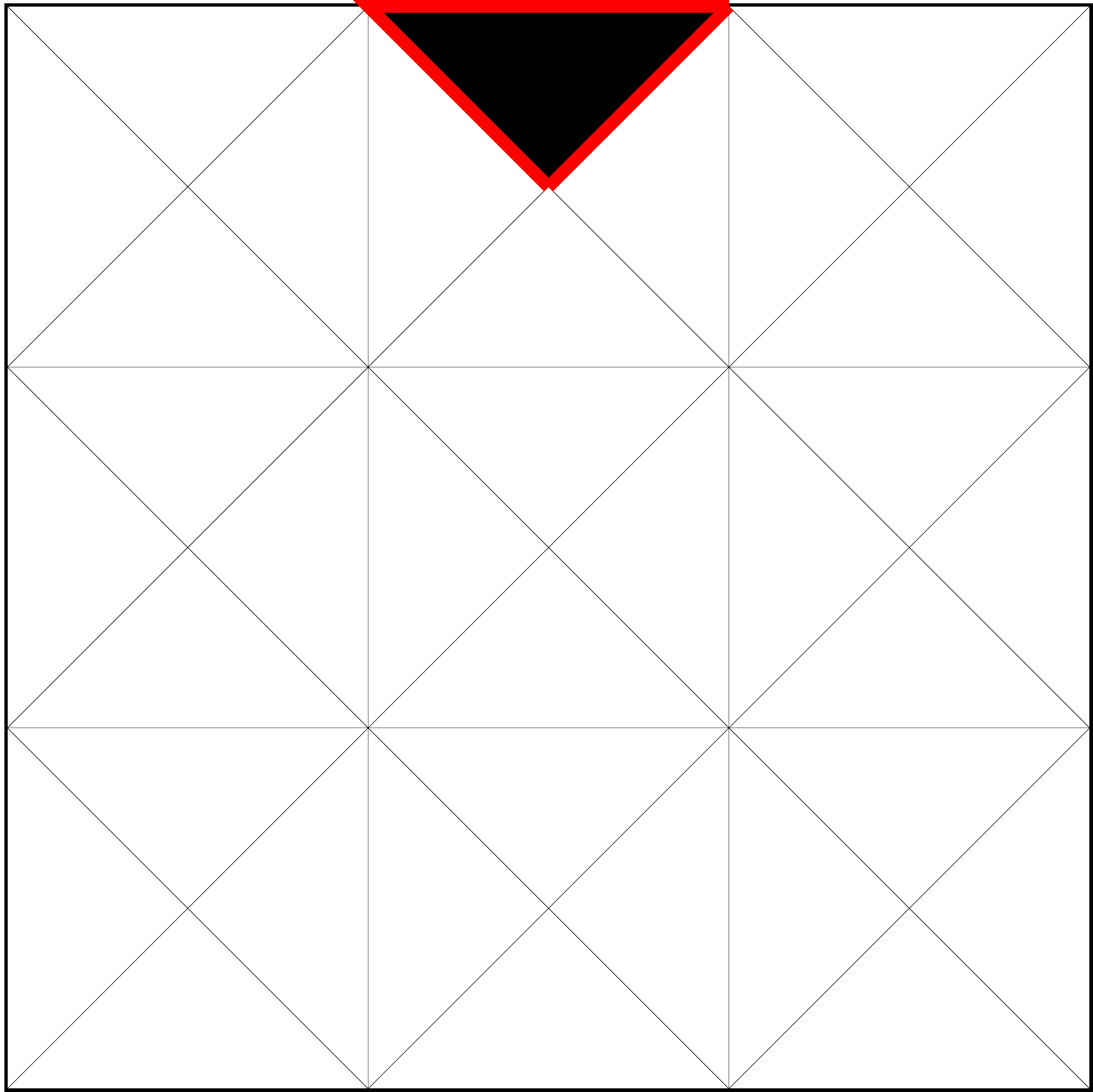}
\caption{Belief. }
\label{experiments:fig:kit:step4:b}
\end{subfigure}

\vspace{1em}

\begin{subfigure}{0.46\columnwidth}
\centering
\includegraphics[width=\columnwidth, trim=0cm 0cm 0cm 0cm, clip]{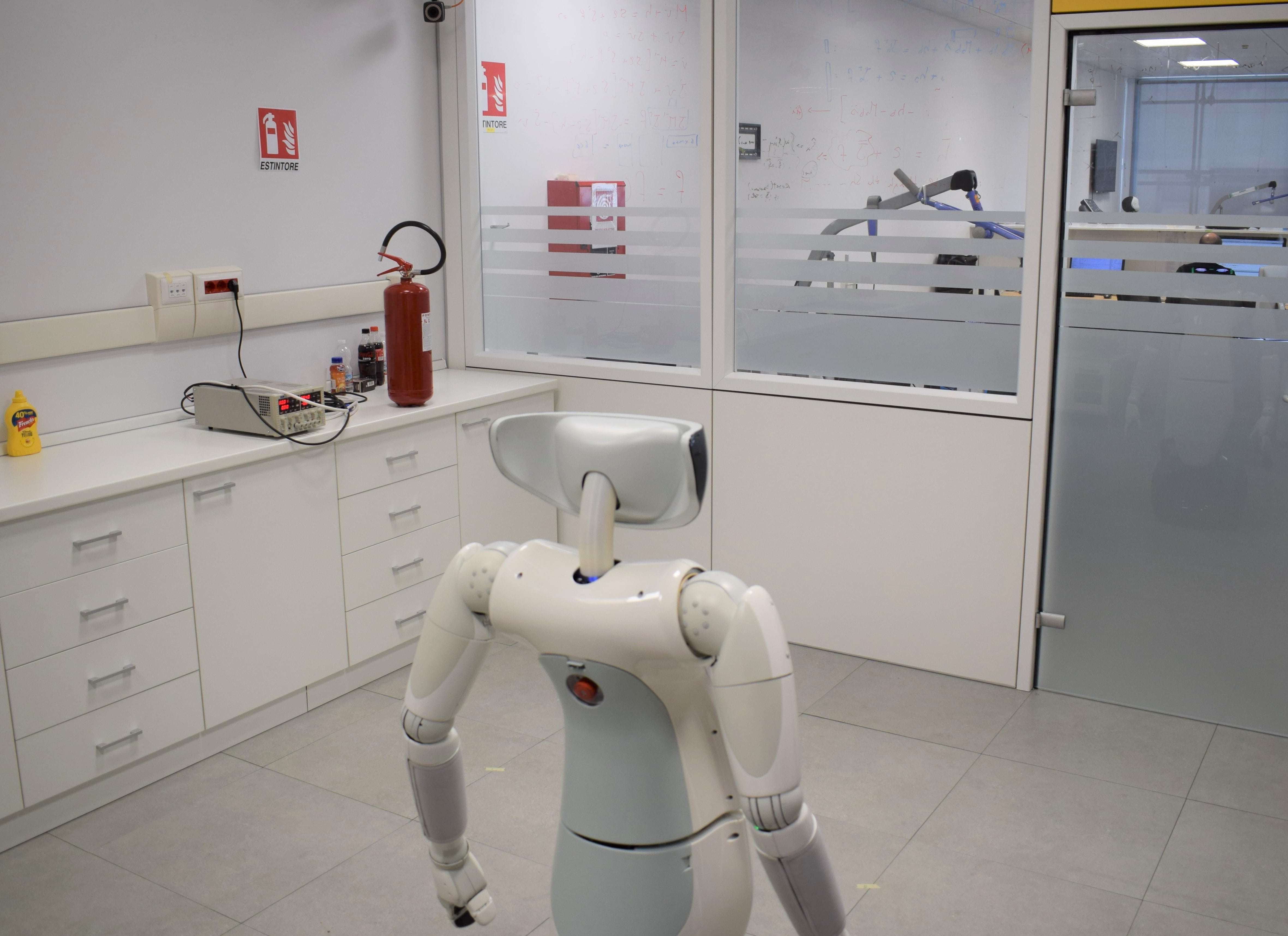}
\caption{Initial (unknown) pose.}
\label{experiments:fig:kit:step1:f}
\end{subfigure}
\begin{subfigure}{0.46\columnwidth}
\centering
\includegraphics[width=\columnwidth, trim=0cm 0cm 0cm 0cm, clip]{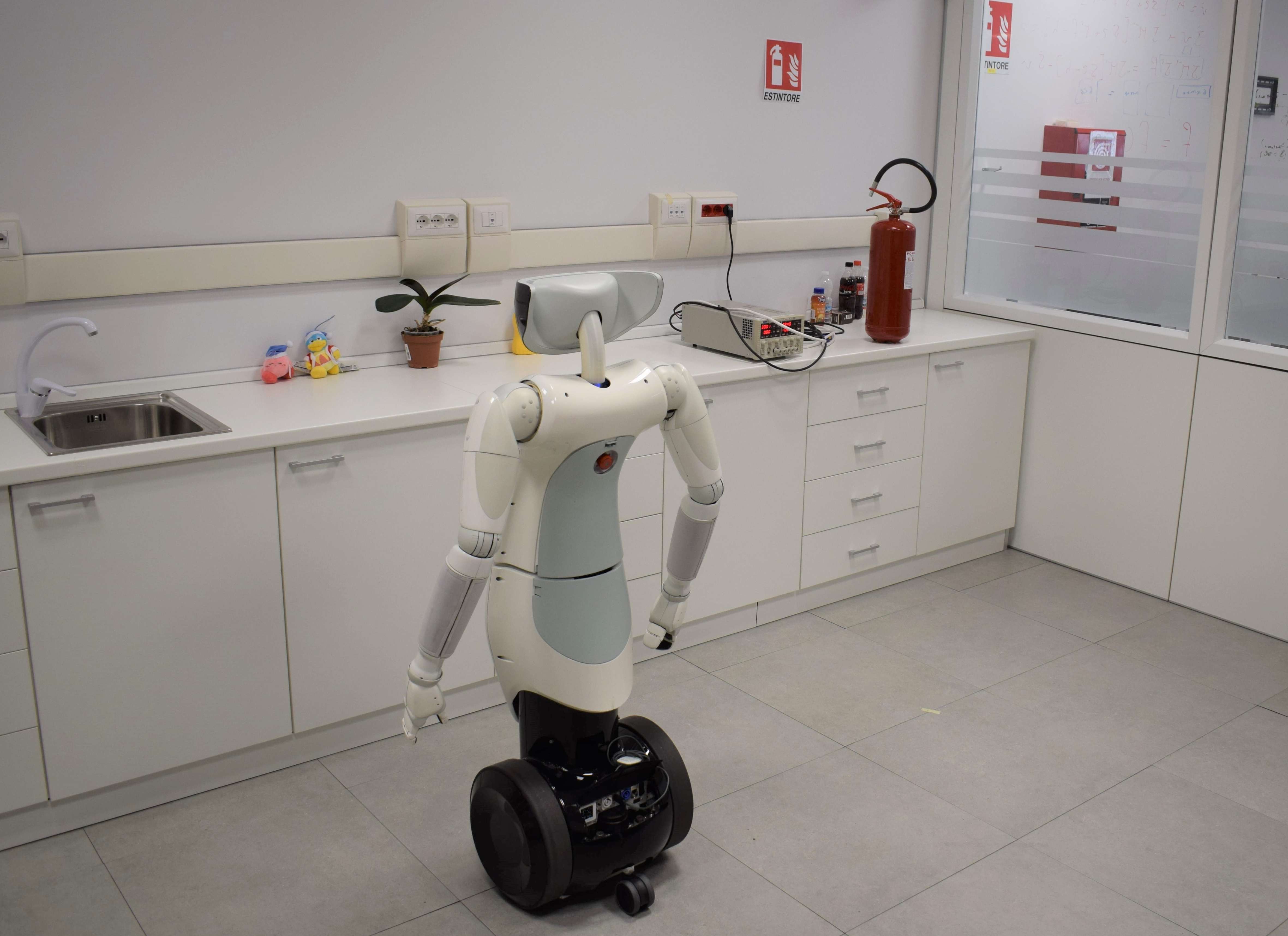}
\caption{Located Pose.}
\label{experiments:fig:kit:step2:f}
\end{subfigure}

\vspace{1em}
\begin{subfigure}{0.082\columnwidth}
\centering
\includegraphics[width=\columnwidth]{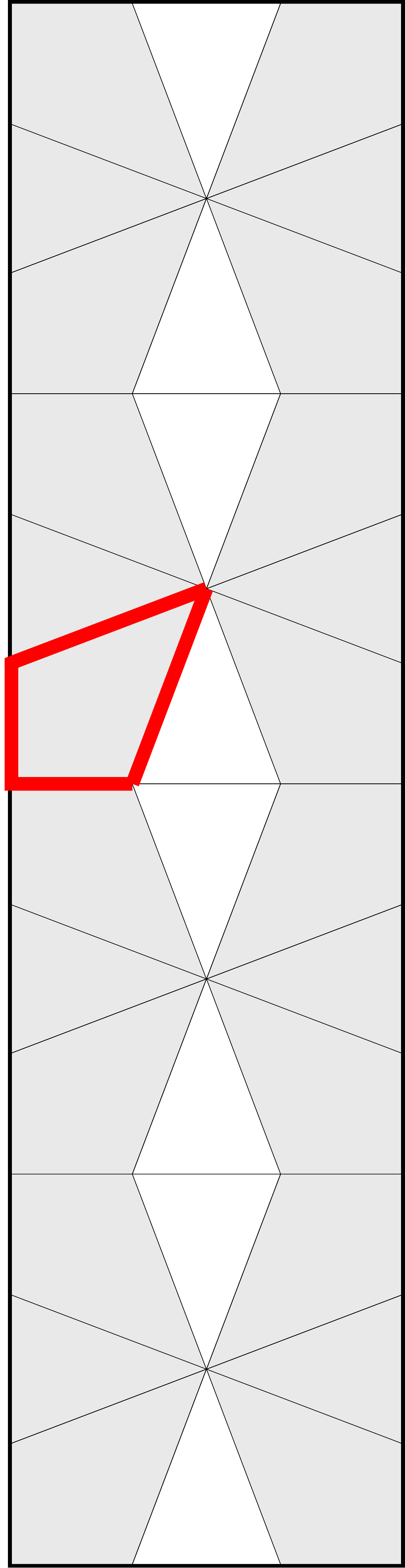}
\caption{B. }
\label{experiments:fig:hal:step0:b}
\end{subfigure}
\begin{subfigure}{0.23\columnwidth}
\centering
\includegraphics[width=\columnwidth, trim=0cm 2cm 0cm 0cm, clip]{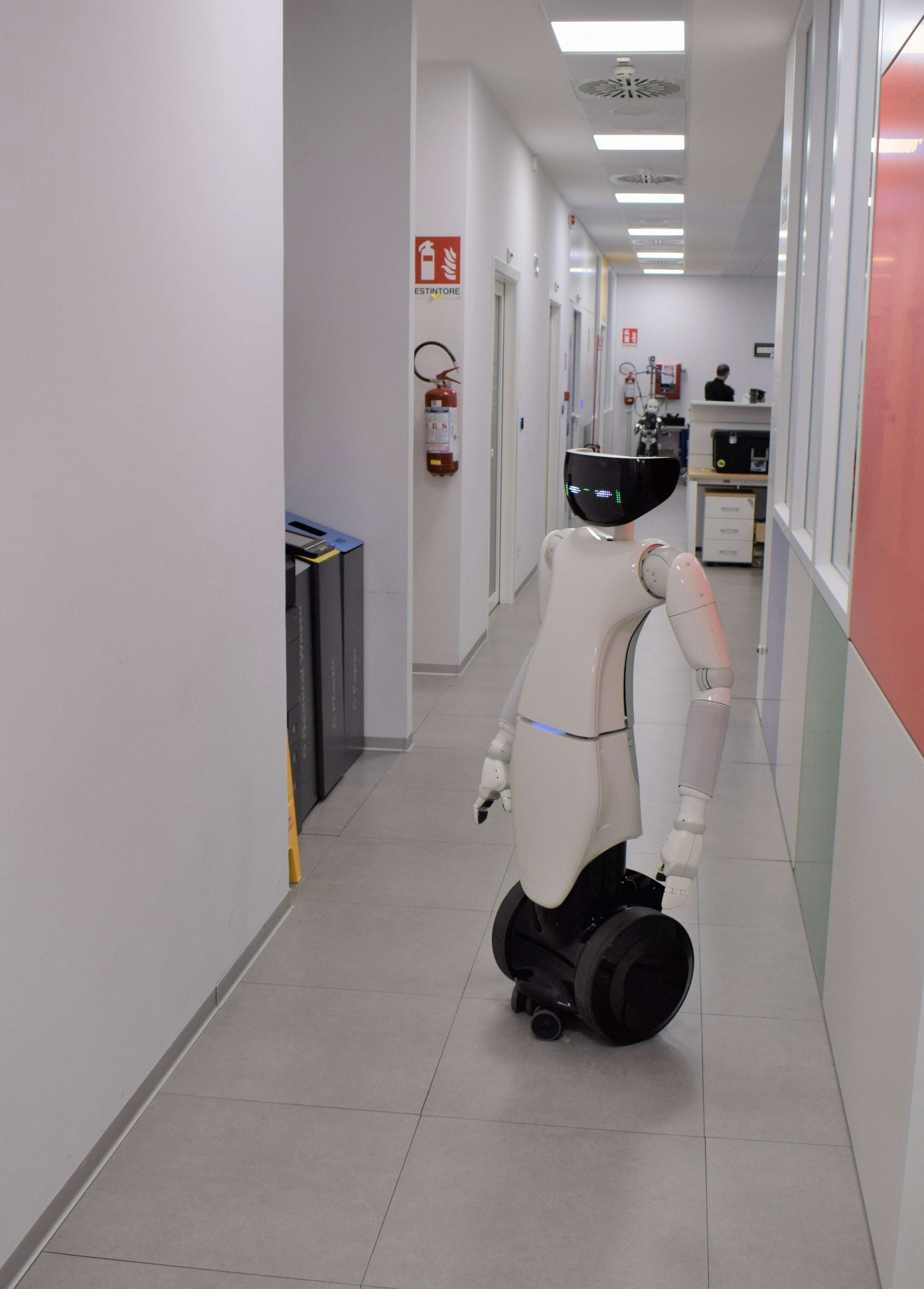}
\caption{Initial Pose. }
\label{experiments:fig:hal:step0:f}
\end{subfigure}
\begin{subfigure}{0.082\columnwidth}
\centering
\includegraphics[width=\columnwidth]{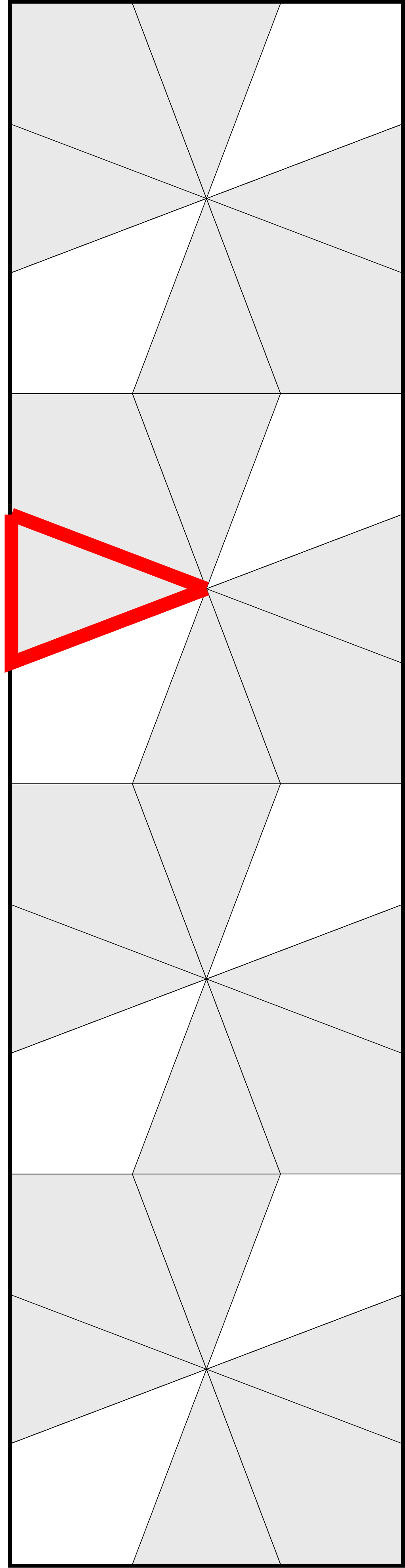}
\caption{B. }
\label{experiments:fig:hal:step1:b}
\end{subfigure}
\begin{subfigure}{0.23\columnwidth}
\centering
\includegraphics[width=\columnwidth, trim=0cm 2cm 0cm 0cm, clip]{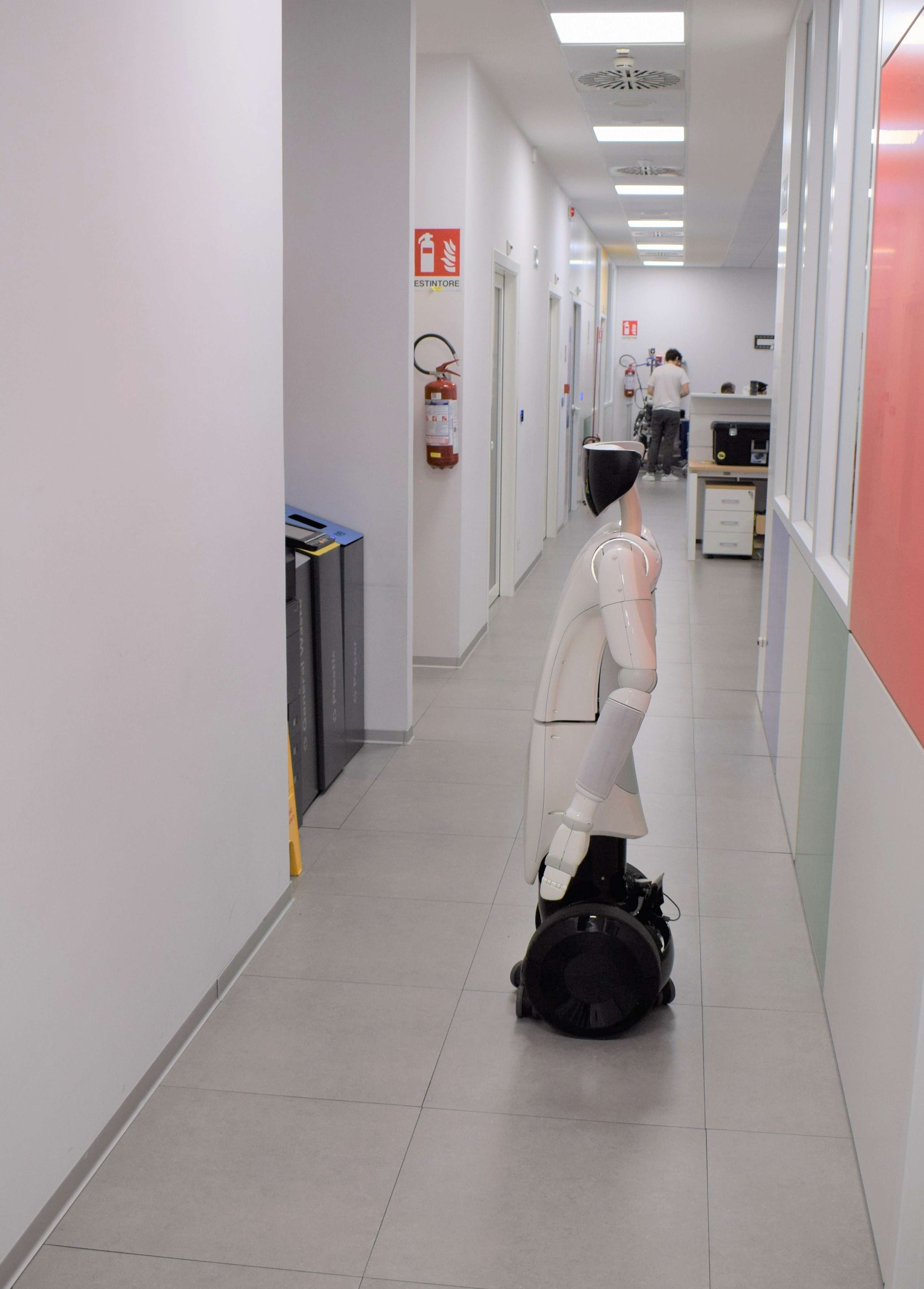}
\caption{Rotating.}
\label{experiments:fig:hal:step1:f}
\end{subfigure}
\begin{subfigure}{0.082\columnwidth}
\centering
\includegraphics[width=\columnwidth]{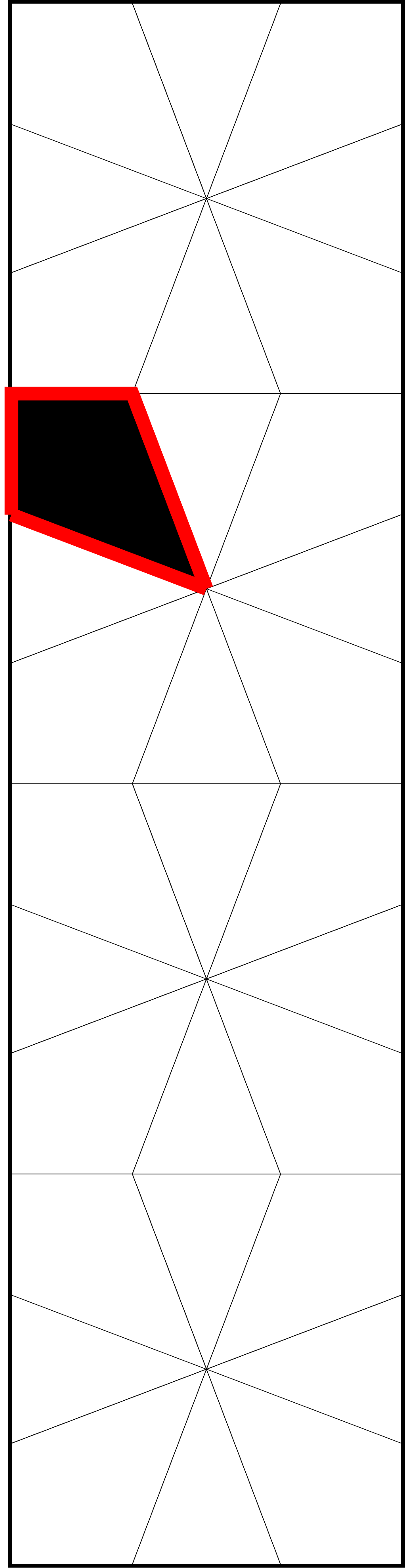}
\caption{B. }
\label{experiments:fig:hal:step2:b}
\end{subfigure}
\begin{subfigure}{0.23\columnwidth}
\centering
\includegraphics[width=\columnwidth, trim=0cm 2cm 0cm 0cm, clip]{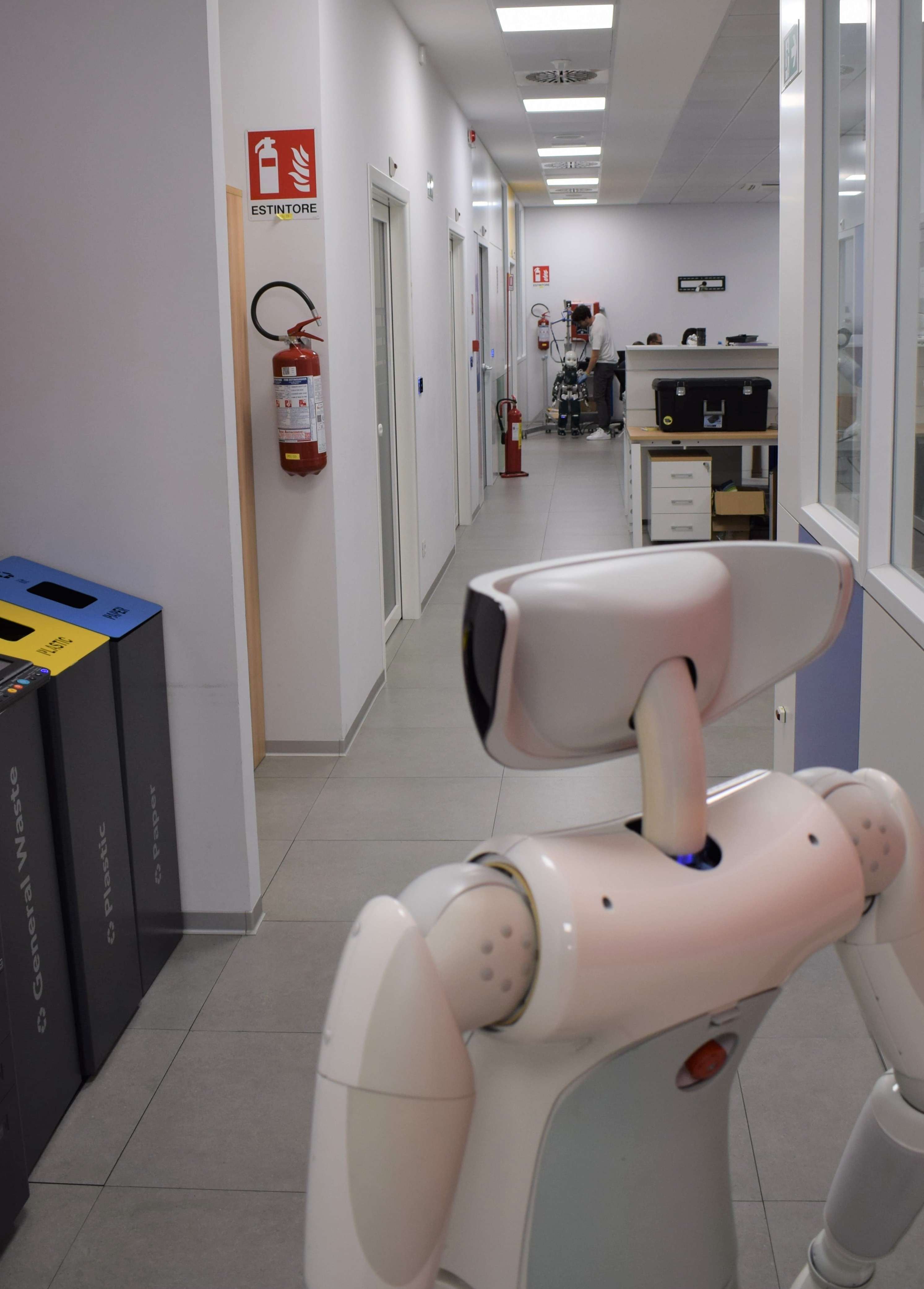}
\caption{Located.}
\label{experiments:fig:hal:step2:f}
\end{subfigure}
\caption{Executions steps (pictures) and belief state representations (grids) of the experiments. Different shades of gray indicate the probability of being inside that triangle. The red triangle represents the current physical state.}
\end{figure}

The environment is surrounded by walls. For simplicity, we consider as wall any surface orthogonal to the floor.
The perception action \emph{Scan} computes, using the laser scanner, the distance to the wall in front, if any. 
Without loss of generality, we consider five possible observations for the action \emph{Scan}: \emph{No Wall}, \emph{Wall closer that 0.2m}, \emph{Wall between 0.2m and 0.8m}, \emph{Wall between 0.8m and 1.1m}, and \emph{Wall between 1.1m and 1.5m}.
The actuation actions are the same of the experiments in Section~\ref{sec:experiments:sim}. 
We employ a low-level local navigation system to compute the relative motion primitives to move the robot. If a motion primitive intersects a wall detected by the scanner, the robot does not move. We construct the matrix $T$ and  vector $O$ as in Section~\ref{sec:experiments:sim} with the exception for the objects label and the additional perception action \emph{Scan}.
We manually constructed a semantically-labeled map of the environment with the objects classes seen and the wall distances detected from the laser scanner at each state. 
\begin{experiment}[Kitchen]
\label{ex:kitchen}
We mapped the kitchen in a 3x3 grid, where each  cell has size 1.2m and it is divided into four triangles.
We placed the robot in the center of the kitchen and ran the experiment. (Figure~\ref{experiments:fig:kit:step1:f}). The initial belief state is the one representing no prior information about the robot's pose (i.e. a vector $b\in \mathbb{R}^{36}$ with each entry equal to $\frac{1}{36}$). The robot first performs the action \emph{Look}, seeing no objects, and updates its belief state to one that represents high probability of being in a physical state from which no objects are seen (Figure~\ref{experiments:fig:kit:step0:b}). Then, it performs the action \emph{Scan}, detecting no walls in front, and updates its belief state accordingly (Figure~\ref{experiments:fig:kit:step1:b}). Then, it performs the action \emph{RotateCounterClockwise} and updates its belief state accordingly (Figure~\ref{experiments:fig:kit:step2:b}). Then, it performs the action  \emph{Look}, seeing only the potted plant, and updates its belief state accordingly (Figure~\ref{experiments:fig:kit:step3:b}). There exist two possible physical states that match, with high probability, the observations and actions computed heretofore. To discriminate them, the robot performs the action \emph{MoveForward} (Figure~\ref{experiments:fig:kit:step2:f}) and then the action  \emph{Scan}, detecting a wall in front between 0.2 and 0.8m. The robot locates itself correctly (Figure~\ref{experiments:fig:kit:step4:b}). 

\end{experiment}
\newpage
\vspace*{-2em}
\begin{experiment}[Hallway]
We mapped the hallway in a 1x4 grid, where each cell has size 1.5m and it is divided into eight triangles. The hallway has a fire extinguisher to one of its ends. We enriched the observations for the action \emph{Look} by including the \emph{viewing angle}, i.e. the angle from the optical axis coming out from the robot's head, of the object classes seen. We used the same setup of Experiment~\ref{ex:kitchen} with the exception that the robot rotates by $\pi/4$ (the cells are divided in eight triangles) and the observations to account for viewing angles.
We placed the robot in the hallway and ran the experiment  (Figure~\ref{experiments:fig:hal:step0:f}). The robot first performs the action \emph{Scan}, detecting a wall, and updates its belief state to one that represents high probability of being in a physical state with a wall in front (Figure~\ref{experiments:fig:hal:step0:b}). Then the robot performs the action 
\emph{RotateClockWise} twice (Figures~\ref{experiments:fig:hal:step1:f} and~\ref{experiments:fig:hal:step2:f}) and updates its belief state accordingly (Figure~\ref{experiments:fig:hal:step1:b} and~\ref{experiments:fig:hal:step2:b}). The robot now performs the action \emph{Look}, seeing a fire extinguisher with a viewing angle between 18 and 30 degrees, and updates its belief state accordingly. There is only one physical state matching the actions and observations above with the desired probability. The robot get correctly located (Figure~\ref{experiments:fig:hal:step2:b}). 
\end{experiment}

%

\section{Concluding Remarks}
We outlined a novel interleaved act and plan framework to obtain a quick approximate of a robot pose exploiting semantic observations and an annotated map. 
Manual annotation is a tedious and error-prone task that limited our real world experiments within a coarse-grained maps, despite our approach handles large state spaces. We are working on automatic annotation using existing approaches \cite{gay2018visual} to get fine-grained maps and obtain higher resolutions.   

\clearpage
\balance
\bibliographystyle{IEEEtran}
\bibliography{refs}
\end{document}